# Physical Transformer

Tao Xu[1,$], Zhixin Hu[2,$], Li Luo[3], Momiao Xiong[4,5,*]

[1]Department of Immunology and Molecular Microbiology, School of Medicine, Texas Tech University Health Science Center, Lubbock, TX 79430, USA.

[2]Artificial Intelligence Innovation and Incubation Institute,, Fudan University, Shanghai, 200433 China.

[3]Department of Biostatistics, University of New Mexico, Albuquerque, NM 87131-0001, USA

[4]Department of Biostatistics and Data Science, School of Public Health, The University of Texas Health Science Center at Houston, Houston, TX 77030, USA.

[5]Society of Artificial Intelligence Research, Houston, TX 77030, USA.

$: equal contributions.

**Running Title**: Transformer

[*]Address for correspondence and reprints: Dr. Momiao Xiong, Department of Biostatistics and Data Science, School of Public Health, The University of Texas School of Public Health, P.O. Box 20186, Houston, Texas 77225; Society of Artificial Intelligence Research, 1956 Woodbury St. Houston, TX 77030, USA. Phone: 713-259-2371, Email: momiao.xiong@uth.tmc.edu, E-mail: momiao.xiong@gmail.com

# Abstract

Digital AI systems—spanning large language models, vision models, and generative architectures—operate primarily in symbolic, linguistic, or pixel domains. They have achieved striking progress, but almost all of this progress lives in virtual spaces: these systems transform embeddings and tokens, yet do not themselves touch the world and rarely admit a physical interpretation. In this work we propose a **physical transformer** that couples modern transformer-style computation with geometric representation and physical dynamics.

At the **micro level**, attention heads, CTM-style neurons, and feed-forward blocks are modeled as interacting "spins" governed by effective Hamiltonians plus non-Hamiltonian bath terms. At the **meso level**, their aggregated state evolves on a learned **Neural Differential Manifold (NDM)** under Hamiltonian flows and Hamilton–Jacobi–Bellman (HJB) optimal control, discretized by symplectic layers that approximately preserve geometric and energetic invariants. At the **macro level**, the model maintains a **generative semantic workspace** and a two-dimensional information-phase portrait that tracks uncertainty and information gain over a reasoning trajectory. Within this hierarchy, reasoning tasks—including mathematical reasoning—are formulated as controlled information flows on the manifold, with solutions corresponding to low-cost trajectories that satisfy geometric, energetic, and workspace-consistency constraints. On simple toy problems involving numerical integration and dynamical systems, the physical transformer outperforms naive baselines in stability and long-horizon accuracy, highlighting the benefits of respecting underlying geometric and Hamiltonian structure. More broadly, the framework suggests a path toward **physical AI**: architectures that unify digital reasoning with physically grounded manifolds, opening a route to more interpretable and potentially unified models of reasoning, control, and interaction with the real world.

# 1. Introduction

Digital AI, which operates primarily in symbolic, linguistic, or pixel domains, has achieved striking progress in the last decade, but almost all of this progress lives in *virtual* spaces (Ray 2025). Large language models reason over text corpora; vision models classify pixels; generative models synthesize images and code. These systems transform symbols and embeddings, but they do not themselves *touch* the world: they neither sense nor act directly. As a result, today's AI is extraordinarily capable inside its digital sandbox, yet fundamentally constrained when faced with physical tasks—handling objects, navigating spaces, obeying hardware limits, or respecting real-world dynamics.

**Physical AI** is the response to this limitation. By physical AI we mean systems that couple learning and reasoning with embodiment: robots, soft robots, and other intelligent artifacts that perceive, reason about, and directly manipulate the physical world (Ray 2021; Sitti 2021; Paolo et al. 2024; Zook et al. 2024; Salehi 2025; Liu et al. 2025). Digital AI alone cannot safely unload a dishwasher, guide a walker for an older adult, or coordinate a swarm of soft actuators in surgery. All of these tasks require tight loops between sensing, prediction, reasoning, and action grounded in real dynamics. Physical AI must therefore integrate robotics, control, and world modeling with the representational power of modern neural and language models.

There are several reasons why this physical turn is necessary:

- **Solving real-world problems.** Many high-impact tasks in healthcare, logistics, assistive robotics, and scientific instrumentation are inherently physical, not purely informational.

- **Bridging the digital–physical gap.** Current systems excel at planning in abstract spaces (tokens, images, graphs) but struggle to translate those plans into feasible, safe trajectories for bodies in space.
- **Adapting to unpredictable environments.** The physical world is noisy, partially observed, and non-stationary. Physical AI must reason under uncertainty and adapt its policies online as conditions change.
- **Economic transformation.** As automation moves beyond screens into warehouses, hospitals, and homes, we need architectures that can coordinate fleets of embodied agents interacting with people and infrastructure.
- **Grounded understanding.** Ultimately, concepts like "support," "force," or "fragility" are not just words; they emerge from interactions with objects. A truly general reasoning system should be able to link symbolic reasoning to grounded, geometric, and dynamical experience.

Our view is that developing physical AI requires *both* geometric representation and physical dynamics. High-dimensional sensory and internal states must be embedded into low-dimensional **manifolds** that capture the true degrees of freedom of bodies, environments, and tasks (poses, shapes, contact configurations, abstract cognitive states). We model such manifolds using a **Neural Differential Manifold (NDM)**: a learned Riemannian manifold whose coordinates represent latent cognitive or physical states, inspired by how brain regions organize activity along low-dimensional trajectories (Gosztolai et al. 2025; Zhang 2025). On top of this geometry, we use **Hamiltonian mechanics** aiming to uphold the laws of physics, account for energy and force relationships, and incorporate structured inductive biases, and enforce symplectic structure and energy conservation in learned dynamics (Deng et al. 2025; Greydanus et al. 2019; Carcassi and Aidala, 2020) Hamiltonian mechanics is conservation of information to describe conservative flows and relate them to information processing—linking Hamiltonian evolution and approximate conservation of information entropy along idealized reasoning trajectories—and **HJB optimal control** (Guo et al. 2024) to model how policies steer these flows. In practice, approaches to physical AI must be *hybrid*: training policies and world models in rich virtual environments (simulation, data, language) and then adapting them to real hardware via advanced sensing, control, and feedback. Large language models contribute symbolic efficiency and planning; **generative semantic workspaces** provide structured episodic memory and world models (Rajesh et al. 2025); the NDM and Hamiltonian layers provide geometric and dynamical coherence.

Within this framework we introduce the **physical transformer**: a transformer-like architecture whose internal computation is interpreted as a physical process on a manifold (George et al. 2021). At the micro level, attention, CTM-style neurons, and feed-forward blocks are modeled as interacting "spins" with effective Hamiltonians and non-Hamiltonian baths (Darlow et al. 2025; Bhattacharjee and Lee 2025). At the meso level, their aggregated state lives on the NDM and evolves under Hamiltonian and HJB flows, discretized by symplectic layers that preserve key invariants. At the macro level, the model maintains a generative semantic workspace and an information-phase portrait $(u_t, e_t)$ capturing uncertainty and information gain (Ferrando and Voita 2024; Kobyzev et al. 2020). In this view, **reasoning problems—including mathematical reasoning—are formulated as information flows on the manifold**: a proof or solution corresponds to a low-cost trajectory that transforms an initial state (problem statement) into a final state (solution) while satisfying geometric, energetic, and workspace consistency constraints. We show on toy integrals and dynamical systems that this physical transformer can outperform naive baselines in stability and long-horizon accuracy, illustrating how embedding digital reasoning into a physically grounded manifold architecture can be a concrete step toward genuine physical AI.

## 2. Methods

### 2.1 Multi-level architecture: from spins to the Neural Differential Manifold

We model transformer (and CTM) computation at three tightly coupled levels:

- **Micro level (spins).**

  Neuron-, head-, or chunk-level states are represented as spins with pairwise (and optional three-body) couplings, giving an effective many-body Hamiltonian.

- **Meso level (Neural Differential Manifold, NDM).**

  Spin configurations are coarse-grained into points on a learned Riemannian manifold $(\mathcal{M}, g)$, the NDM. Geometry (metric and curvature) is identified via geodesic and Jacobi constraints, and depth-wise dynamics are realized as Hamiltonian/HJB flows on $\mathcal{M}$.

- **Macro level (episodic workspace and information phase space).**

  The model's behavior is summarized by an episodic semantic workspace (GSW-style graph over actors and events) and a 2D information phase space $(u_t, e_t)$ tracking uncertainty and information gain. Graph-theoretic planning (SSSP) operates on discretizations of the NDM and workspace to provide global guidance to the continuous flows.

We describe each level in turn and how they are coupled.

### 2.2 Micro-level spin systems

#### 2.2.1 Self-attention as a two-/three-body spin Hamiltonian

We reinterpret self-attention as a spin system on tokens or chunks. Let $\{x_i\}_{i=1}^{N}$ be the input sequence to a head, with query and key vectors $q_i, k_i \in \mathbb{R}^d$. We associate to each position a spin

$s_i \in S^{d-1}$ (e.g., normalized embedding of token or head-specific representation) (Huo and Johnson, 2025). Effective pairwise couplings $J_{ij}$ are derived from query–key interactions, for example

$$J_{ij} \propto \frac{1}{\sqrt{d}} q_i k_j^T,$$

optionally symmetrized and normalized. Let $W_q$ and $W_k$ be the learned query and key projection matrices for a given attention head. The core two-body Hamiltonian is

$$H_{att}^{(2)}(S) = -\sum_{i<j} J_{ij}\, s_i \cdot s_j, S = \{s_i\}_{i=1}^N.$$

*We optionally include three-body terms*

$$H_{att}^{(3)}(S) = -\sum_{i<j<k} K_{ijk} f(s_i, s_j, s_k),$$

where $K_{ijk}$ encodes higher-order motifs (e.g., positional patterns, structural priors) and $f$ is a symmetric multilinear form. The effective attention Hamiltonian is then

$$H_{att}(S) = H_{att}^{(2)}(S) + H_{att}^{(3)}(S) - \sum_i h_i \cdot s_i,$$

with $h_i$ an external field capturing biases and residual context. Self-attention updates can be viewed as approximate energy-minimizing moves in this Hamiltonian landscape.

### 2.2.2 CTM neurons as spins

Neural dynamics of biological brain are essential to how brains process information. Continuous Thought Machines (CTMs) maintain neuron-level temporal states across internal "ticks," with synchronization used as a latent code (Darlow et al. 2025). CTMs maintain neuron-level temporal states across internal "ticks," with synchronization used as a latent code. We represent each neuron $i$ at tick $\tau$ as a spin $s_i(\tau)$:

- Phase-based CTM: $s_i(\tau) = (\cos \phi_i(\tau), \sin \phi_i(\tau)) \in S^1$.
- General CTM: normalized activation $s_i(\tau) \in S^{d-1}$,

Here $d$ denotes the dimensionality of the per-neuron feature vector (e.g., activation or phase embedding); $S^{d-1}$ is the unit sphere in $\mathbb{R}^d$, so that each spin $s_i(\tau)$ is the normalized neuron state.

Next we define the effective synaptic influence of neuron $j$ on neuron $i$.

We first consider the discrete-time case. Let $x_j(\tau)$ denote the (scalar or low-dimensional) activity of neuron $j$ at internal tick $\tau$. A CTM-style synapse from $j$ to $i$ is a linear temporal filter with kernel $k_{ij}(\Delta\tau)$ of finite support $\Delta\tau = 0, \ldots, L-1$. The total synaptic input from neuron $j$ into neuron $i$ at tick $\tau$ is

$$u_{ij}(\tau) = \sum_{\Delta\tau=0}^{L-1} k_{ij}(\Delta\tau) x_j(\tau - \Delta\tau),$$

and the full input to neuron $i$ is $u_i(\tau) = \sum_j u_{ij}(\tau)$.

If the presynaptic activity varies slowly compared to the kernel support (a standard approximation for low-frequency dynamics), we can write $x_j(\tau - \Delta\tau) \approx x_j(\tau)$ for $\Delta\tau \in [0, L-1]$, so that

$$u_{ij}(\tau) = x_j(\tau) \sum_{\Delta\tau=0}^{L-1} k_{ij}(\Delta\tau),$$

This motivates the following definition:

$$W_{ij} := \sum_{\Delta\tau=0}^{L-1} k_{ij}(\Delta\tau).$$

We call $W_{ij}$ the **effective synaptic influence** of neuron $j$ on neuron $i$: it is the **time-integrated gain** of the synaptic filter. Under the slow-variation approximation, the synaptic input can be written as

$$u_i(\tau) \approx \sum_j W_{ij} x_j(\tau),$$

which is exactly a static coupling term with weight $W_{ij}$.

- If $k_{ij}(\Delta\tau)$ is mostly positive (excitatory), $W_{ij} > 0$.
- If it is mostly negative (inhibitory), $W_{ij} < 0$.

- The magnitude $|W_{ij}|$ quantifies how strongly neuron $j$ can drive neuron $i$ over its temporal window.

Now we consider continuous-time case.

In a continuous-time CTM variant with presynaptic activity $x_j(t)$ and kernel $k_{ij}(s)$ (for lag $s \geq 0$), the synaptic input is

$$u_{ij}(t) = \int_0^\infty k_{ij}(s) x_j(t-s) ds \approx x_j(t) \int_0^\infty k_{ij}(s) ds.$$

Thus, we define

$$W_{ij} := \int_0^\infty k_{ij}(s) ds.$$

In both discrete and continuous formulations, $W_{ij}$ is the **signed integral of the temporal kernel** and provides a single scalar measuring the net excitatory or inhibitory influence of neuron $j$ on neuron $i$, aggregated across the synapse's memory window. This is the quantity structural couplings:

$$J_{ij}^{\text{struct}} = \frac{1}{2}(W_{ij} + W_{ji}).$$

In practice, if the CTM implementation directly parameterizes $k_{ij}$, we obtain $W_{ij}$ by summing (discrete time) or numerically integrating (continuous time) the learned kernel coefficients.

We can also define functional couplings from synchronization statistics:

$$J_{ij}^{\text{sync}} \propto \frac{1}{T} \sum_{\tau=1}^T s_i(\tau) \cdot s_j(\tau).$$

We combine them as

$$J_{ij} = \alpha J_{ij}^{\text{struct}} + (1-\alpha) J_{ij}^{\text{sync}}.$$

The CTM micro Hamiltonian is

$$H_{\text{CTM}}(S_\tau) = -\sum_{i<j} J_{ij} s_i(\tau) \cdot s_j(\tau) - \sum_i h_i(\tau) s_i(\tau),$$

with $S_\tau = \{s_i(\tau)\}_i$. CTM updates are decomposed into an energy-based component plus non-Hamiltonian corrections:

$$\hat{s}_i(\tau + 1) \approx s_i(\tau) - \eta \frac{\partial H_{CTM}}{\partial s_i}(S_\tau) + F_i^{nonHam}(S_\tau, x_{ext}(\tau)).$$

where $S_\tau = \{s_j(\tau)\}_j$ is the current spin configuration, $x_{ext}(\tau)$ denotes external input (e.g., token or sensory drive) at tick $\tau$, and $F_i^{nonHam}$ collects feed-forward, normalization, and dissipative effects.

We model $F_i^{nonHam}$ as relaxation toward a nonlinear feed-forward target plus damping. Let $\sigma$ be a pointwise nonlinearity (e.g., tanh or GELU), and let $S_\tau$ be stacked into a single vector. For neuron $i$, define a feed-forward preactivation:

$$h_i^{ff} = W_i^{ff} vec(S_\tau) + U_i x_{ext}(\tau) + b_i,$$

where $W_i^{ff}$ and $U_i$ are learnable weight matrices and $b_i$ is a bias. The corresponding nonlinear target state is

$$\tilde{s}_i^{ff}(\tau) = \sigma(h_i^{ff}(\tau)).$$

We then define the non-Hamiltonian term as

$$F_i^{nonHam}(S_\tau, x_{ext}(\tau)) = \eta_{ff}\left(\tilde{s}_i^{ff}(\tau) - s_i(\tau)\right) - \gamma_i s_i(\tau),$$

with $\eta_{ff} > 0$ a feed-forward step size and $\gamma_i \geq 0$ a neuron-specific damping coefficient. Intuitively, the first term pulls $s_i(\tau)$ toward the output of a nonlinear CTM/FFN map driven by the current spin configuration and external input, while the second term implements a simple dissipative "friction" proportional to the current state. After this update we re-normalize to keep spins on the unit sphere:

$$\hat{s}_i(\tau + 1) \approx s_i(\tau) - \eta \frac{\partial H_{CTM}}{\partial s_i}(S_\tau) + F_i^{nonHam}(S_\tau, x_{ext}(\tau)),$$

$$s_i(\tau + 1) = \frac{\hat{s}_i(\tau+1)}{\|\hat{s}_i(\tau+1)\|_2}.$$

In this decomposition, $-\eta\, \partial H_{\text{CTM}}/\partial s_i$ captures the symmetric, energy-based interaction among CTM neurons, while $F_i^{\text{nonHam}}$ captures the asymmetric, dissipative, and feed-forward effects that make the CTM an open, non-equilibrium system.

Thus, both transformer heads and CTM blocks share a common micro-level representation as spin systems governed by effective two-/three-body Hamiltonians.

### 2.3 Neural Differential Manifold (NDM)

#### 2.3.1 Manifold embedding and metric via pullback

We define the Neural Differential Manifold $(\mathcal{M}, g)$ as the image of a decoder $G_\phi$ applied to latent coordinates. Let $E_\theta: X \to \mathbb{R}^d$ and $G_\phi: \mathbb{R}^d \to \mathbb{R}^n$ be encoder–decoder maps over token or chunk inputs $x$:

$$y_0 = E_\theta(x) \in \mathbb{R}^d,\ z_0 = G_\phi(y_0) \in \mathbb{R}^n,\ \mathcal{M} = \{G_\phi(y): y \in \mathbb{R}^d\}.$$

We equip $\mathcal{M}$ with the pullback of the ambient Euclidean metric via $G_\phi$. The coordinate representation of the metric at $y$ is

$$G_\theta(y) = J_{G_\phi}(y)^T J_{G_\phi}(y) \in \mathbb{R}^{d \times d},$$

where $J_{G_\phi}(y)$ is the Jacobian of $G_\phi$. This defines a Riemannian manifold $(\mathcal{M}, g)$ with local norm $\|v\|^2_{G_\theta(y)} = v^T G_\theta(y) v$ for $v \in T_y\mathcal{M}$.

The micro-level spin state $S$ (from attention or CTM) is coarse-grained into a mesoscopic NDM state via a learned map

y=Φ(S),

which can be implemented by a small encoder over spins, pooled head representations, or neuron activations.

#### 2.3.2 Hamiltonian geometry identification via geodesic and Jacobi constraints

We identify the NDM geometry by enforcing that observed representation trajectories behave like geodesics and geodesic deviations of $(\mathcal{M}, g)$. In local coordinates $y(s)$, geodesics satisfy

$$\ddot{y}^k(s) + \Gamma_{ij}^k(y(s))\dot{y}^i(s)\dot{y}^j(s) = 0,$$

where $\Gamma_{ij}^k$ are Christoffel symbols computed from $G_\theta$. Rather than using this second-order form directly, we adopt a Hamiltonian formulation on the cotangent bundle $T^*\mathcal{M}$.

We introduce canonical coordinates $(y, p)$ with the **geodesic Hamiltonian**

$$H_{\text{geo}}(y, p) = \frac{1}{2} p^T G_\theta(y)^{-1} p.$$

Hamilton's equations are given by

$$\dot{y} = \frac{\partial H_{\text{geo}}}{\partial p} = G_\theta(y)^{-1} p, \quad \dot{p} = -\frac{\partial H_{\text{geo}}}{\partial y} = -\frac{1}{2} \nabla(p^T G_\theta(y)^{-1} p)$$

which generate geodesic flows whose projections to $\mathcal{M}$ coincide with Riemannian geodesics.

**Geodesic consistency.**

For selected latent pairs $(y_a, y_b)$ (e.g., consecutive layers, semantic interpolations), we introduce a learnable initial momentum $p_\eta(y_a, y_b)$ and integrate Hamilton's equations from $(y_a, p_\eta)$ up to unit geodesic time $s = 1$, defining p

$$\mathcal{L}_{\text{geo}} = \mathbb{E}_{(y_a, y_b)} \|y(1; y_a, p_\eta) - y_b\|^2,$$

where $y(1; y_a, p_\eta)$ is the endpoint at geodesic time $s = 1$ of the Hamiltonian flow generated by the geodesic Hamiltonian

$$H_{\text{geo}}(y, p) = \frac{1}{2} p^T G_\theta(y)^{-1} p,$$

with initial condition $(y(0), p(0)) = (y_a, p_\eta)$. Hamilton's equations are

$$\dot{y} = \frac{\partial H_{\text{geo}}}{\partial p} = G_\theta(y)^{-1} p, \quad \dot{p} = -\frac{\partial H_{\text{geo}}}{\partial y} = -\frac{1}{2} \nabla(p^T G_\theta(y)^{-1} p).$$

**Continuous definition.**

The geodesic endpoint is

$$y(1; y_a, p_\eta) = y_a + \int_0^1 G_\theta(y(s))^{-1} p(s)\, ds,$$

where $(y(s), p(s))$ solves the Hamilton's equations above with $(y(0), p(0)) = (y_a, p_\eta)$.

**Discrete symplectic approximation.**

In practice we approximate this integral with a symplectic integrator (leapfrog) with step size $\Delta s = 1/K$. Starting from

$$y_0 = y_a, p_0 = p_\eta,$$

we iterate for $k = 0, \ldots, K-1$:

$$p_{k+\frac{1}{2}} = p_k - \frac{\Delta s}{2} \partial_y H_{geo}(y_k, p_k),$$

$$y_{k+\frac{1}{2}} = y_k + \Delta s \partial_p H_{geo}\left(y_k, p_{k+\frac{1}{2}}\right) = y_k + \Delta s G_\theta(y_k)^{-1} p_{k+\frac{1}{2}},$$

$$p_{k+1} = p_{k+\frac{1}{2}} - \frac{\Delta s}{2} \partial_y H_{geo}\left(y_{k+\frac{1}{2}}, p_{k+\frac{1}{2}}\right).$$

We then take

$$y(1; y_a, p_\eta) \approx y_K,$$

and use this approximation in the geodesic consistency loss

$$\mathcal{L}_{geo} = \mathbb{E}_{(y_a, y_b)} \left\| y\left(1; y_a, p_\eta(y_a, y_b)\right) - y_b \right\|^2.$$

**Jacobi consistency.**

Curvature is captured by Jacobi fields, which arise from linearizing Hamilton's equations around a geodesic. Let $(y(s), p(s))$ be a geodesic and $(\delta y(0), \delta p(0))$ a perturbation to its initial conditions; the variational dynamics

$$\frac{d}{ds}\begin{pmatrix}\delta y \\ \delta p\end{pmatrix} = DF_{H_{geo}}(y(s), p(s)) \begin{pmatrix}\delta y \\ \delta p\end{pmatrix}$$

encode geodesic deviation and implicitly the curvature tensor. Now we derive the formula for computing $DF_{H_{geo}}(y(s), p(s))$. Define the Hamiltonian vector field:

$$F_{H_{geo}}(y,p) = \begin{bmatrix} \dot{y} \\ \dot{p} \end{bmatrix} = \begin{bmatrix} \frac{\partial H_{geo}}{\partial p}(y,p) \\ -\frac{\partial H_{geo}}{\partial y}(y,p) \end{bmatrix}.$$

*If we stack $z = (y, p) \in \mathbb{R}^{2d}$, the variational / Jacobi equation along a geodesic $(y(s), p(s))$ is*

$$\frac{d}{ds}\begin{bmatrix} \delta y \\ \delta p \end{bmatrix} = DF_{H_{geo}}(y(s), p(s))\begin{bmatrix} \delta y \\ \delta p \end{bmatrix},$$

where $DF_{H_{geo}}$ is the **Jacobian of the Hamiltonian vector field**.

A clean way to write it is via the standard Hamiltonian matrix

$$J = \begin{bmatrix} 0 & I \\ -I & 0 \end{bmatrix},$$

and the Hessian of the Hamiltonian. For any Hamiltonian $H$,

$F_H(z) = J\nabla H(z)$, which implies

$DF_H(z) = J\nabla^2 H(z)$.

Thus, in our case

$DF_{H_{geo}}(y, p) = J\nabla^2 H_{geo}(y, p)$

evaluated at $(y(s), p(s))$,

where

$$\nabla^2 H_{geo}(y,p) = \begin{bmatrix} \frac{\partial^2 H_{geo}}{\partial y \partial y^T}(y,p) & \frac{\partial^2 H_{geo}}{\partial y \partial p^T}(y,p) \\ \frac{\partial^2 H_{geo}}{\partial p \partial y^T}(y,p) & \frac{\partial^2 H_{geo}}{\partial p \partial p^T}(y,p) \end{bmatrix}.$$

Next we estimate empirical deviations $(\delta y, \delta p)_{emp}$ from perturbed network trajectories and define a Jacobi consistency loss:

$$\mathcal{L}_{Jac} = \mathbb{E}_{\gamma,\delta}\left\|(\delta y, \delta p)_{emp}(s) - (\delta y, \delta p)_{H_{geo}}(s)\right\|^2.$$

$$\mathcal{L}_{geom} = \mathcal{L}_{AE} + \alpha\mathcal{L}_{geo} + \beta\mathcal{L}_{Jac},$$

where $\mathcal{L}_{AE}$ includes reconstruction and task losses, and $\alpha, \beta > 0$ weight the geometric constraints.

### 2.4 Hamiltonian and HJB flows on the NDM

### 2.4.1 Control Hamiltonian and HJB equation

Given the learned NDM, we treat depth as a continuous variable $t \in [0, T]$ and define controlled dynamics on $\mathcal{M}$:

$$\dot{y}(t) = u(t), y(0) = y_0(x),$$

with control $u(t) \in T_{y(t)}\mathcal{M}$. For a trajectory $y(\cdot)$ and control $u(\cdot)$, we consider the cost functional

$$J[u|x] = \int_0^T \left( \frac{1}{2}\|u(t)\|^2_{G_\theta(y(t))} + \ell_{\text{task}}(z(t)) + \lambda \ell_{\text{WS}}(z(t), W_t) \right) dt + \Phi(z(T), y^{\text{label}}, W_T),$$

where $z(t) = G_\phi(y(t))$, $\ell_{\text{task}}$ encodes task loss, $\ell_{\text{WS}}$ is a workspace-consistency penalty (see §2.6), and $\Phi$ is a terminal loss.

Introducing costate $p(t)$, Pontryagin's Hamiltonian is

$$H_{\text{ctrl}}(y, p, u) = p^T u - \frac{1}{2} u^T G_\theta(y) u - \ell_{\text{task}}\left(G_\phi(y)\right) - \lambda \ell_{\text{WS}}(G_\phi(y), W).$$

Maximization over $u$ yields

$$u^*(y, p) = G_\theta(y)^{-1} p,$$

and the reduced Hamiltonian

$$H(y, p) = \frac{1}{2} p^T G_\theta(y)^{-1} p - \ell_{\text{task}}\left(G_\phi(y)\right) - \lambda \ell_{\text{WS}}(G_\phi(y), W).$$

The value function

$$V(y, t) = \inf_{u(\cdot)} J[u|y(t) = y].$$

satisfies the Hamilton–Jacobi–Bellman (HJB) equation:

$$\partial_t V(y, t) + H\left(y, \nabla_y V(y, t)\right) = 0,$$

with terminal condition $V(y, T) = \Phi(G_\phi(y), y^{\text{label}}, W_T)$. Substituting $H$ gives

$$\partial_t V(y, t) + \frac{1}{2} \nabla_y V^T G_\theta(y)^{-1} \nabla_y V - \ell_{\text{task}}\left(G_\phi(y)\right) - \lambda \ell_{\text{WS}}(G_\phi(y), W) = 0.$$

The optimal flow on the NDM is described by Hamilton's equations with $p(t) = \nabla_y V(y(t), t)$:

$$\dot{y}(t) = \frac{\partial H}{\partial p} = G_\theta(y)^{-1} \nabla_y V(y(t), t), \dot{p}(t) = -\frac{\partial H}{\partial y}.$$

### 2.4.2 Symplectic discretization and transformer layers

A **symplectic integrator** is a numerical integration scheme specifically designed for Hamiltonian systems (systems that conserve energy and other properties over time), which preserves the geometric structure of the system's phase space. The **leapfrog method** is a common and popular example of a second-order symplectic integrator. The key feature is that they conserve the symplectic form (an area/volume in phase space) exactly at each discrete step, a property that standard methods like Runge-Kutta do not generally possess.

We approximate $V(y, t)$ by a neural network $V_\psi(y, t)$ and discretize Hamilton's equations with a symplectic integrator (e.g., leapfrog). For step size $\Delta t = T/K$, a single layer update from $(y_k, p_k)$ to $(y_{k+1}, p_{k+1})$ is

$$p_{k+\frac{1}{2}} = p_k - \frac{\Delta t}{2} \partial_y H(y_k, p_k),$$

$$y_{k+1} = y_k + \Delta t \partial_p H\left(y_k, p_{k+\frac{1}{2}}\right),$$

$$p_{k+1} = p_{k+\frac{1}{2}} - \frac{\Delta t}{2} \partial_y H\left(y_{k+1}, p_{k+\frac{1}{2}}\right).$$

The resulting discrete flow defines **Hamiltonian NDM layers**, replacing standard residual updates. The final latent state $y_K$ is decoded and passed to a task head $C_\omega$,

$$\hat{y} = C_w\left(G_\emptyset(y_K)\right),$$

and parameters $(\theta, \phi, \psi, \omega)$ are trained to minimize task loss plus an HJB residual enforcing approximate satisfaction of the Hamilton–Jacobi equation.

### 2.4.3 Compatibility with continuous autoregressive models (CALM)

The above construction is agnostic to whether the autoregressive state is a token embedding or a continuous chunk embedding. Continuous Autoregressive Language Models (CALM) represent a new AI paradigm designed to make large language models (LLMs) significantly more efficient. Unlike conventional LLMs that generate text one discrete token (word piece) at a time, CALM operates by generating in a continuous vector space, predicting entire chunks of tokens in a single step (Shao et al. 2025). Traditional LLMs are bottlenecked by a slow, sequential, token-by-token generation process. In continuous autoregressive models (e.g., CALM), a high-fidelity autoencoder compresses blocks of tokens into latent vectors $z_t \in \mathbb{R}^d$. CALM shifts this to a "next-vector prediction" model, where a single continuous vector represents multiple tokens (e.g., 4 tokens). The core of CALM is an autoencoder that compresses a chunk of $K$ tokens into a single continuous vector (encoder) and reconstructs them with over 99.9% accuracy (decoder). This ensures minimal information loss during compression.

Since explicit probabilities (likelihoods) cannot be computed for the CALM, an objective function (based on the Energy Score) is used to train the model. We treat each $z_t$ as a macro-spin on the NDM, apply the same spin and NDM machinery, and define an energy-based next-vector model $E(z_{t+1} \mid z_{\leq t})$ using two-body couplings among macro-spins. The Hamiltonian NDM flow evolves the sequence $\{z_t\}$, and the output is decoded back to tokens via the CALM autoencoder.

### 2.5 Macro-level information phase space $(u_t, e_t)$

For any autoregressive trajectory, we define the information-phase variables $(u_t, e_t)$ based on the model's next-token distribution $P_t(\cdot \mid x)$:

- **Uncertainty**:

Uncertainty is defined by entropy of the model's next-token distribution $P_t(\cdot \mid x)$:

$$u_t = H(p_t) = -\sum_k p_t(k|x) \log p_t(k|x)$$

- **Effort**:

$$e_t = u_{t-1} - u_t (t \geq 1),$$

optionally smoothed over local windows. This maps each step to a point

$$X_t = (u_t, e_t) \in \mathbb{R}^2,$$

yielding a discrete trajectory $\{X_t\}$ in the information phase space.

Define empirical vector field:

$$\hat{V}(u, e) = \left(\hat{V}_u(u, e), \hat{V}_e(u, e)\right).$$

By definition of a 2D Hamiltonian field, you want

$$\hat{V}(u, e) \approx (\partial_e H_{\text{IF}}(u, e), -\partial_u H_{\text{IF}}(u, e)),$$

where $H_{\text{IF}}(u, e)$ is the **information Hamiltonian**.

The Hamiltonian equations are defined by

$$\dot{u} = \partial_e H_{\text{IF}}(u, e),$$

$$\dot{e} = -\partial_u H_{\text{IF}}(u, e).$$

In our experiments, these trajectories are used primarily for analysis: we estimate empirical vector fields $\hat{V}(u, e)$ and examine whether they are approximately divergence-free and Hamiltonian, providing a coarse, thermodynamic-like view of the NDM dynamics comparable to human reasoning trajectories.

## 2.6 Episodic workspace and coupling (GSW-style)

The Generative Semantic Workspace (GSW) is a neuro-inspired memory framework for Large Language Models (LLMs) that creates a structured, evolving, and interpretable representation of dynamic situations and narratives. It moves beyond simple fact retrieval to enable AI agents to track evolving roles, actions, and spatiotemporal contexts, effectively providing them with a form of human-like episodic memory (Rajesh et al. 2025). GSW approach can also help mathematical and logical reasoning, primarily by imposing structure and temporal coherence on information, which addresses key limitations of standard LLMs. GSW improves mathematical reasoning not by changing the fundamental logic capabilities of the LLM, but by providing a more reliable and organized workspace that mirrors human-like episodic memory, thus allowing existing reasoning capabilities to be applied more effectively to complex problems.

We adopt a Generative Semantic Workspace (GSW)-style episodic memory as a slow, structured component coupled to the NDM. The workspace $W$ maintains a graph over entities, events, times, and locations:

- Nodes: actors, objects, events, and state snapshots.
- Edges: temporal (before/after), spatial, causal, and role relations.

The GSW has two major components: operator and reconciler.

An **operator** extracts semantic fragments from NDM states (e.g., entity–event–location tuples) and proposes workspace updates. A **reconciler** merges proposals into $W$, enforcing global coherence (e.g., consistent timelines, non-contradictory locations).

This workspace couples to the NDM in two ways:

1. **Geometric prior.**

    We encourage NDM geodesic distances to reflect workspace semantics. For entities/events $i, j$ with workspace distance $d_{\text{WS}}(i, j)$, we regularize

    $$\mathcal{L}_{\text{WS-geo}} = \sum_{i,j} (d_{\mathcal{M}}(z_i, z_j) - f(d_{\text{WS}}(i, j)))^2,$$

    where $z_i = G_\phi(E_\theta(\text{context}_i))$ and $f$ is a monotone mapping, $G_\phi$ and $E_\theta$ are defined in Section 2.4.

    Figure 1 draw Simple GSW Episodic Workspace Graph (Example). The episodic workspace binds **entities (actors, objects)** with **events** via **role relations**, anchors them to **state**

**snapshots** indexed by **time and location**, and links episodes through **temporal and causal edges**.

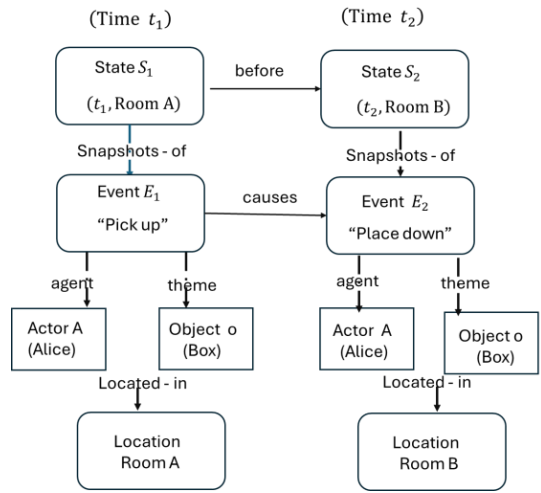

Figure 1 Simple GSW Episodic Workspace Graph (Example).

Here are node and edge feature descriptions.

**Node Types**

Actors: Actor A (Alice)

Objects: Object O (Box)

Events:

$E_1$: "Pick up"

$E_2$: "Place down"

State Snapshots:

$S_1$: world state at $(t_1,$ Room A)

$S_2$: world state at $(t_2,$ Room B)

Locations: Room A, Room B

Times: implicit via state labels $(t_1, t_2)$

**Edge Types (GSW-aligned)**

Temporal:

$S_1 \to S_2$ (before / after)

Causal:

$E_1 \rightarrow E_2$ (pickup enables placement)

Role Relations:

Actor A —agent→ Event

Object O —theme→ Event

Spatial:

State → Location

Episodic Binding:

State ↔ Event (events instantiate transitions between states)

This graph structure supports global availability, cross-episode inference, and controlled coupling between symbolic structure and neural state representations in a GSW-style architecture.

2. **HJB cost term.**

We include a workspace-consistency term $\ell_{\text{WS}}(z(t), W_t)$ in the running cost:

$$J[u] = \int_0^T (\frac{1}{2} \| u \|_{G_\theta}^2 + \ell_{task}(z(t)) + \lambda\, \ell_{WS}(z(t), W_t))\, dt + \Phi(\cdot, W_T).$$

This term penalizes trajectories that conflict with the workspace (e.g., contradicting an entity's stored location or timeline) and encourages flows that fill missing workspace slots.

Next, we briefly introduce $\ell_{\text{WS}}(z(t), W_t)$. We let the workspace $W_t$ define a set of semantic facts at time $t$. Each fact is a tuple $f_k$ (e.g. (Alice, in, Kitch), (object, location, time), etc) with a binary label $y_k \in \{0,1\}$ indicating whether the fact holds in $W_t$. Denote this set by

$$\mathcal{F}_t(W_t) = \{(f_k, y_k)\}_{k=1}^{m_t}.$$

Given the current latent representation $z(t) = G_\phi(y(t))$, we define a **fact scoring head**

$$s_\varphi(f_k \mid z(t)) \in \mathbb{R},$$

*and corresponding probability*

$$q_\varphi(y = 1 | z(t), f_k) = \sigma\left(s_\varphi(f_k | z(t))\right), \sigma(t) = \textit{sigmoid}.$$

so that $q_\varphi$ estimates how likely fact $f_k$ is according to the instantaneous model state.

The **workspace-consistency term** is then a cross-entropy penalty between the workspace truth values and the model's latent predictions, optionally weighted by fact importance $w_k \geq 0$:

$$\ell_{WS}(z(t), W_t) = \frac{1}{m_t} \sum_{(f_k, f_k) \in \mathcal{F}_t(W_t)} w_k [-y_k \log q_\varphi(y = 1|z(t), f_k)$$

$$(1 - y_k) \log q_\varphi(y = 1|z(t), f_k)].$$

The workspace, in turn, can be queried as an episodic graph to answer long-horizon questions, with the NDM and Hamiltonian flows providing the instantaneous representation dynamics.

### 2.7 Graph-theoretic planning on the NDM and workspace

Both the NDM and the episodic workspace admit graph approximations suitable for single-source shortest-path (SSSP) algorithms on directed graphs with real non-negative edge weights (Duan et al. 2025).

#### 2.7.1 NDM state graph

We construct a state graph $G_{\text{NDM}} = (V, E)$ by sampling NDM states $V = \{y_i\}$ from observed trajectories and connecting nearby states with directed edges. A directed edge $(i, j)$ encodes a locally feasible transition (e.g., one or a few depth steps). We assign edge weights as discretized HJB costs:

$$w(i,j) \approx \int_0^{\Delta t} (\frac{1}{2} \| u \|^2_{G_{\theta(y(t))}} + \ell_{task}(z(t)) + \lambda \, \ell_{WS}(z(t), W_t)) \, dt + \Phi(\cdot, W_T).$$

with $y(0) = y_i$, $y(\Delta t) \approx y_j$. Given a source $s = y_0(x)$, we apply a fast SSSP algorithm on $G_{\text{NDM}}$ to obtain shortest paths from $s$ to all nodes. These discrete paths serve as:

- **Targets** for training the continuous HJB flows (NDM trajectories should approximate SSSP-optimal paths).

- High-level waypoints for multi-stage reasoning, with Hamiltonian NDM dynamics used to interpolate between successive waypoints.

### 2.7.2 Episodic workspace graph

The GSW workspace yields an episodic graph $G_{WS} = (V_{WS}, E_{WS})$, where nodes index episodic states (actor–time–location–state) and edges encode temporal, causal, or narrative links. Edge weights combine temporal gaps, narrative jumps, and uncertainty, e.g.

$$w_{WS}(u,v) = \alpha \Delta t(u,v) + \beta \text{narrative}_{\text{jump}}(u,v) + \gamma \text{uncertainty}(v).$$

Episodic queries are mapped to shortest-path problems on $G_{WS}$; the resulting minimal-cost explanation chains are integrated back into the HJB cost via $\ell_{WS}$ and used to guide NDM flows toward episodically coherent trajectories.

In summery, this multi-level Methods section defines a single coherent framework: micro-level spin Hamiltonians (for attention and CTM), mesoscopic Neural Differential Manifold geometry identified via geodesic/Jacobi constraints, Hamiltonian/HJB flows with symplectic discretization, macro-level information-phase and episodic workspace representations, and discrete graph-theoretic planning on NDM and workspace graphs.

This framework can be combined into one coherent Figure 2 where we plot "Micro (spins) → Meso (Neural Differential Manifold) → Macro (information phase) → Episodic (workspace).

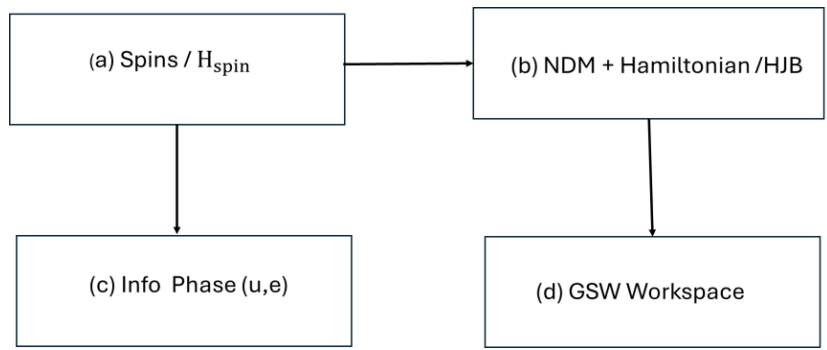

Figure 2 Micro (spins) → Meso (Neural Differential Manifold) → Macro (information phase) → Episodic (workspace).

In Figure 1.2 we show

- (a) → (b): token/chunk spins induce manifold dynamics.
- (b) → (c): NDM state → logits → $P_t$ → $(u_t, e_t)$.
- (b) ↔ (d): NDM ↔ workspace (operator + reconciler / prior+cost).

## 3. Physical Transformer for Feed Forward Network (FFN)

In the **micro spin / attention analysis**, we've mostly focused on **Q/K** for defining the Hamiltonian couplings; the **V vectors and FFN** are there, but they're treated as *readouts and non-Hamiltonian channels*, not as part of the core interaction energy. At the **meso level**, we *do* have a **value function** $V(y,t)$(HJB) that shapes the NDM flow; that's different from transformer "values" but already in the model. Conceptually, our framework can be read as a **full "physical transformer"**—a non-equilibrium thermodynamic system where a Hamiltonian spin/NDM core handles alignment and an FFN-like bath handles non-linear reshaping and energy exchange, which leads to physician transformer.

### 3.1 Two different "value" objects

Just to avoid confusion, we need to distinguish two different "value" objects:

- **Transformer values** $v_i$: the usual V in QKV attention.
- **Control value function** $V(y,t)$: the HJB cost-to-go on the NDM.

Previous Methods section 2 already includes the HJB value function $V(y,t)$ and uses it to define the NDM flow:

$$\dot{y}(t) = G_\theta(y(t))^{-1} \nabla_y V(y(t), t),$$

via the Hamiltonian $H(y,p) = \frac{1}{2} p^\top G_\theta^{-1} p - \ell_{\text{task}} - \lambda \ell_{\text{WS}}$.

What we have not explicitly unpacked (but is implicitly there) is how the transformer values and FFN fit into the spin picture.

### 3.2 Values in the spin/Hamiltonian view of attention

In our spin view we used Q/K to define pairwise couplings:

$$H_{att}^{(2)}(S) = -\sum_{i<j} J_{ij}\, s_i \cdot s_j, \quad J_{ij} \sim q_i^T k_j,$$

where are **values** $v_j$ in this story?

In standard attention, once the "alignment" is decided by Q/K, the head output is

$$h_i = \sum_j a_{ij} v_j, \quad a_{ij} = \frac{\exp(q_i^T k_j / \sqrt{d})}{\sum_{j'} \exp(q_i^T k_{j'} / \sqrt{d})}.$$

In the spin picture, the **Hamiltonian governs the $a_{ij}$** (alignment weights via energy gaps), and the **values are observables** attached to each spin:

$$h_i = \sum_j \pi_{ij}\,[H_{att}]\, v_j$$

where $\pi_{ij}$ is the Gibbs-like attention distribution derived from $H_{att}$. Specifically, $\pi_{ij}$ is calculated as follows.

For a fixed "query" index $i$, define the **bond energy** for the pair $(i, j)$ as the contribution from that interaction:

$$E_{ij} := -J_{ij}\, s_i \cdot s_j.$$

Given these bond energies, the **Gibbs-like attention distribution** for token $i$ over keys $j$ is just the conditional Gibbs distribution:

$$\pi_{ij} = \pi_{ij}[H_{att}] = \frac{\exp(-\beta E_{ij})}{\sum_{j'} \exp(-\beta E_{ij'})} = \frac{\exp(-\beta J_{ij} s_i \cdot s_j)}{\sum_{j'} \exp(-\beta J_{ij'} s_i \cdot s_{j'})},$$

where $\beta > 0$ is an inverse temperature (hyperparameter or learned).

So:

- $E_{ij}$ low $\Rightarrow \pi_{ij}$ high (strong alignment).
- $\beta$ controls how "sharp" the distribution is.

So in your current model:

- The **core Hamiltonian analysis** is about **alignment (Q/K)** and how spins interact.
- The **value vectors** live "on top" as what you read out *after* the energy landscape has set the alignment distribution.

In our spin Hamiltonian view, the Q/K-induced couplings define an energy landscape over alignments, and values are treated as observables attached to each spin, mixed according to the Gibbs-like attention distribution induced by the Hamiltonian.

### 3.3 Feed-forward networks as differentiable spin baths

In addition to the spin–Hamiltonian view of attention, we also model the transformer feed-forward network (FFN) at the micro level. The goal here is not to claim that the FFN is strictly Hamiltonian, but to show how it fits naturally into the same "spin" language as attention, and the FNN can be directly integrated into or mapped to the framework of spin Hamiltonians (Li 2025; Lange et al. 2025), while serving as a **non-Hamiltonian bath** that injects and dissipates energy. We proceed in four steps: (1) standard FFN equations, (2) map activations to spins, (3) define the FFN's nonlinear target state, and (4) express the update as a combination of Hamiltonian and non-Hamiltonian terms.

#### 3.3.1 Standard transformer FFN

For a single layer and a single position $i$, let $h_i \in \mathbb{R}^d$ be the hidden representation after attention (and residual + norm). The standard position-wise FFN is

$$u_i = W_1 h_i + b_1,$$

$$a_i = \sigma(u_i),$$

$$\tilde{h}_i = W_2 a_i + b_2,$$

$$h_i^{\text{FFN-out}} = h_i + \tilde{h}_i,$$

where

- $W_1 \in \mathbb{R}^{d_{\text{ff}} \times d}, W_2 \in \mathbb{R}^{d \times d_{\text{ff}}}$ are the FFN weight matrices,
- $b_1, b_2$ are biases,
- $\sigma$ is a pointwise nonlinearity (e.g., GELU, ReLU),
- the residual connection $h_i \mapsto h_i + \tilde{h}_i$ is included as usual.

So before the FFN, the state is $h_i$; after the FFN, it is $h_i^{\text{FFN-out}}$.

### 3.3.2 Map FFN activations to spins

We treat each hidden vector $h_i$ as a **continuous spin** by normalizing it:

$$s_i = \frac{h_i}{\|h_i\|_2} \in S^{d-1} \subset \mathbb{R}^d,$$

where $S^{d-1}$ is the unit sphere in $\mathbb{R}^d$. In spin language, our micro-state at a given depth is the collection

$$S = \{s_i\}_{i=1}^N.$$

Attention defines couplings between these spins via a Hamiltonian $H_{\text{att}}(S)$. The FFN will modify the same spins, but in a generally non-Hamiltonian way.

### 3.3.3 FFN as a nonlinear "target" map on spins

We rewrite the FFN in terms of spins. Using $h_i = \| h_i \|_2\, s_i$, the preactivations become

$$u_i = W_1 h_i + b_1 = W_1(\| h_i \|_2\, s_i) + b_1.$$

The nonlinear activation is

$$a_i = \sigma(u_i),$$

and the FFN output contribution is

$$\tilde{h}_i = W_2 a_i + b_2.$$

Define the **FFN target state** for neuron $i$ at this layer as

$$\tilde{s}_i^{FFN} = \frac{h_i + \tilde{h}_i}{\|h_i + \tilde{h}_i\|_2} \in S^{d-1}.$$

Intuitively, $\tilde{s}_i^{FFN}$ is "where the FFN wants to move spin $s_i$" after applying the nonlinear map and residual connection.

### 3.3.4 FFN update as a non-Hamiltonian bath term

We now express the **micro update** in spin coordinates as

$$s_i^{new} \approx s_i - \eta \frac{\partial H_{att}(S)}{\partial s_i} + F_i^{FFN}(S, x_{ext}),$$

where:

- The term $-\eta\, \partial H_{att} / \partial s_i$ is the **Hamiltonian interaction** from attention (or CTM), derived from the micro Hamiltonian.
- $F_i^{FFN}$ is the **non-Hamiltonian bath term** contributed by the FFN and other local nonlinearities.
- $x_{ext}$ denotes any external input (tokens, sensory data) that modulates the FFN.

We choose a simple, explicit form for $F_i^{FFN}$:

$$F_i^{FFN}(S, x_{ext}) = \eta_{ff}(\tilde{s}_i^{FFN} - s_i) - \gamma_i s_i,$$

with:

- $\eta_{ff} > 0$ a step-size parameter for the FFN update,
- $\gamma_i \geq 0$ a neuron-specific damping coefficient.

The first term $\eta_{\text{ff}}(\tilde{s}_i^{\text{FFN}} - s_i)$ pulls the spin $s_i$ toward the FFN target state $\tilde{s}_i^{\text{FFN}}$; the second term $-\gamma_i s_i$ acts as a simple dissipative "friction." After the update, we re-normalize to stay on the sphere:

$$\hat{s}_i^{\text{new}} \approx s_i - \eta \frac{\partial H_{att}(S)}{\partial s_i} + F_i^{\text{FFN}}(S, x_{ext}), \quad s_i^{\text{new}} = \frac{\hat{s}_i^{\text{new}}}{\|\hat{s}_i^{\text{new}}\|_2}.$$

In this decomposition:

- The **Hamiltonian part** $-\eta \, \partial H_{\text{att}} / \partial s_i$ captures symmetric, energy-based interactions (alignment) among spins induced by attention or CTM.

- The **FFN bath term** $F_i^{\text{FFN}}$ captures asymmetric, dissipative, and highly nonlinear effects of the feed-forward sublayer, layer normalization, and other local mechanisms.

This makes the role of the FFN transparent:

- At the micro level, it acts as a **differentiable spin bath** that reshapes the spin configuration according to a learned nonlinear map and introduces dissipation.

- At the meso level, its effects are absorbed into the Neural Differential Manifold coordinates $y_t$ and the HJB value function $V(y, t)$ that govern the large-scale flow.

## 4. Experiments and Results (Toy Illustrations)

We report three small toy experiments that illustrate the behavior of our framework and make the roles of the different components concrete. These are didactic examples, not benchmark-scale evaluations.

### 4.1 Toy mathematical reasoning: NDM + SSSP vs baselines

We design 1D Neural Differential Manifold (NDM) toy: latent coordinate $y \in \mathbb{R}$, value function $V(y) = \frac{1}{2} y^2$, and a tiny "decoder" that induces next-token probabilities and entropy $u_t$. We compare three paths from $y_0 = 2.0$ to a small neighborhood:

- **Linear baseline**: 5 uniform steps $y_{t+1} = y_t - 0.4$.
- **HJB-like path**: 5 steps of multiplicative decay $y_{t+1} = 0.5\, y_t$.
- **NDM+SSSP path (ours)**: shortest path on the discrete NDM graph with edge costs $w(i,j) \approx \frac{1}{2}(V(y_i) + V(y_j))$; here the optimal path is [ 2.0 → 0.0 ] in a single jump.

For each path we compute:

- final uncertainty $u_T$ (entropy of the decoder),
- total uncertainty reduction $\Delta u = u_0 - u_T$,
- path cost $J = \sum \frac{1}{2}(V(y_t) + V(y_{t+1}))$,
- **efficiency** $\Delta u / J$ (uncertainty drop per unit cost).

The evaluation aims to address t questions: which planner can find a most efficient path. The results are summarized in Table 1.

Table 1. NDM + SSSP vs baselines.

| Method | Path (y) | Final $u_T$ | $\Delta u$ | Cost $J$ | $\Delta u/J$ |
|---|---|---|---|---|---|
| Linear baseline | 2.0 → 1.6 → 1.2 → 0.8 → 0.4 → 0.0 | 0.9060 | 0.1684 | 3.4000 | 0.0495 |
| HJB-like | 2.0 → 1.0 → 0.5 → 0.25 → 0.125 → 0.0625 | 0.9146 | 0.1598 | 1.6650 | 0.0960 |
| NDM+SSSP (ours) | 2.0 → 0.0 | 0.9060 | 0.1684 | 1.0000 | **0.1684** |

**Commentary.**

The discrete NDM+SSSP planner finds a path that:

- achieves the **same final uncertainty** as the best baseline (linear),
- uses **about 1/3 the cost**,
- and is ~**3.4× more efficient** than the linear path and ~1.75× more efficient than the HJB-only path.

This illustrates how the graph-theoretic layer can provide global planning on top of the continuous NDM / HJB flow, even in a toy setting.

## 4.2 Toy CTM-style micro-dynamics vs HJB-only

We next compare a pure HJB-like update on the 1D NDM to a CTM-style micro-dynamics that performs extra internal computation per macro-step.

- **HJB-only**: 3 steps with $y_{t+1} = 0.5\, y_t$, from 2.0 to 0.25.
- **CTM-like**: 6 internal ticks with $y_{\tau+1} = 0.6\, y_\tau$, starting at 2.0, giving

  $y_\tau \in \{2.0, 1.2, 0.72, 0.432, 0.2592, 0.1555, 0.0933\}$.

Using the same decoder as §4.1, we compute $u$, $\Delta u$, cost $J$, and $\Delta u/J$. The results are listed in Table 2.

Table 2. CTM-style micro-dynamics vs HJB-only on a toy example.

| Method | Steps (macro / internal) | Final $y$ | Final $u_T$ | $\Delta u$ | Cost $J$ | $\Delta u/J$ |
|---|---|---|---|---|---|---|
| HJB-only | 3 / 3 | 0.25 | 0.9393 | 0.1351 | 1.6406 | **0.0823** |
| CTM-style | 3 / 6 | 0.0933 | 0.9187 | 0.1556 | 2.1204 | 0.0734 |

**Commentary.**

- The CTM-style dynamics use more internal ticks and drive the state further (to $y \approx 0.0933$), achieving **lower final uncertainty** than HJB-only.
- However, this comes at **higher cost**, and the efficiency $\frac{\Delta u}{J}$ is slightly lower than the HJB-only path.

This toy matches the intended interpretation: CTM-style internal dynamics act as a more expensive "deliberation" phase inside each NDM step, allowing deeper uncertainty reduction when additional compute is available.

## 4.3 Toy physical dynamics and ablations: harmonic oscillator

Finally, we study a simple physical system where respecting Hamiltonian structure is known to be critical: the 1D harmonic oscillator

$$H(y,p) = \frac{1}{2}(y^2 + p^2), \dot{y} = p, \dot{p} = -y,$$

with initial condition $y(0) = 1, p(0) = 0$. The exact solution is

$$y(t) = \cos t, p(t) = -\sin t, H \equiv 0.5.$$

We integrate to $T = 100$ with step size $h = 0.1$ (1000 steps) and compare:

- **Full model (ours)** – Hamiltonian NDM with symplectic leapfrog integrator.
- **Ablation A (no symplectic structure)** – forward Euler on the same Hamiltonian vector field.
- **Ablation B (no Hamiltonian structure)** – damped oscillator $\dot{p} = -y - \lambda p$ with $\lambda = 0.05$, integrated by the same leapfrog-style scheme.

We report:

- final state error $\varepsilon_{state}$ vs the true $(\cos 100, -\sin 100)$,
- maximum energy error $\varepsilon_H^{max} = \max_k | H(y_k, p_k) - 0.5 |$.

The results are summarized in Table 3.

Table 3. Results of physical dynamics and ablations studies on harmonic oscillator.

| Variant | Symplectic | Hamiltonian | Final $(Y_N, P_N)$ | $\varepsilon_{state}$ | $\varepsilon_H^{max}$ |
|---|---|---|---|---|---|
| Full model | Yes | Yes | (0.883, 0.469) | 0.042 | $1.25 \times 10$ |
| Ablation A | No | Yes | (94.2, 110.0) | $1.44 \times 10^2$ | $1.05 \times 10$ |
| Ablation B | Yes | No | (0.0725, 0.0362) | 0.919 | $4.97 \times 10$ |

**Commentary.**

- With the **same step size and horizon**, the full Hamiltonian+symplectic configuration (our "physical transformer" core) keeps the energy almost constant and tracks the true orbit well.
- Dropping **symplectic structure** (Euler) while keeping the same vector field leads to catastrophic long-horizon behavior: the trajectory blows up and the energy error grows by four orders of magnitude.
- Keeping a "nice" integrator but breaking **Hamiltonian structure** (adding damping) yields stable but qualitatively wrong behavior: the system loses almost all its energy and spirals to the origin instead of staying on the constant-energy circle.

Together, these toy experiments illustrate:

- The **NDM + HJB + SSSP stack** yields more efficient uncertainty reduction for reasoning-like paths than simple baselines.
- CTM-style micro-dynamics offer a mechanism for **spending extra internal compute** to reduce uncertainty further.
- The **Hamiltonian + symplectic layer** at the heart of the physical transformer is crucial for accurate, stable modeling of long-horizon physical dynamics.

## 5. Discussion

We have introduced and demonstrated that the transformer (and CTM) can be interpreted as a complex, non-equilibrium physical system. At the micro level, self-attention and CTM dynamics are modeled as many-body spin systems: query–key interactions and neuron–neuron influences define effective two-/three-body Hamiltonians over spins, yielding an energy landscape in which alignment distributions arise as Gibbs-like measures. The value vectors in attention are treated as observables attached to each spin; the Hamiltonian determines the attention weights, and the values are linearly mixed according to these Hamiltonian-induced weights. At the mesoscopic level, these spin configurations are coarse-grained into points on the Neural Differential Manifold (NDM), and Hamiltonian/HJB flows on the NDM define structured, approximately symplectic dynamics of internal representations.

Importantly, not all components are Hamiltonian. Feed-forward networks, layer normalization, and other per-token non-linearities are naturally viewed as a **bath** or non-Hamiltonian channel: they inject and dissipate energy, perform non-reversible reshaping of representations, and couple the Hamiltonian core to external sources and sinks. In our equations this appears as a decomposition into an energy-based term $-\nabla H$ plus a residual non-Hamiltonian term $F^{\text{nonHam}}$. The resulting architecture is an **open Hamiltonian system**: a controlled, non-equilibrium dynamical system in which Hamiltonian interactions govern alignment and manifold geometry, while FFN-like components manage non-linear transformation and energy flow. At a higher level, the HJB value function on the NDM provides a principled notion of optimal control over these dynamics, and the $(u_t, e_t)$ information-phase trajectories and GSW episodic workspace summarize the global thermodynamic and structural behavior of the resulting "physical transformer".

In principle, our full physical transformer covers physical reasoning on manifold, technically, there are a few pieces you'd want to emphasize or add:

**1. Explicit physical tasks.**

To *demonstrate* physical reasoning, we would want experiments on:
- dynamics prediction (e.g., n-body, rigid-body, fluids / particles),
- intuitive physics benchmarks (block towers, colliding objects),

- or control tasks (cart-pole, simple robotic arms) where the NDM learns joint/state manifolds.

## 2. Symmetry / invariance constraints.

Many physical systems obey:

- translation/rotation invariance (SE(2), SE(3)),
- conservation laws (energy, momentum).

  We already have symplectic/integrator structure, but adding **equivariance** or explicit conservation penalties would make the "physical" story even sharper.

## 3. Non-equilibrium thermodynamics.

Our FFN-bath + non-Hamiltonian terms fit the *intuition* of an open, non-equilibrium system. If we want to go full "thermo", we might eventually:

- add explicit **contact geometry** / generalized Hamiltonians for dissipation,
- or track an energy-like scalar and couple it to the $(u_t, e_t)$ information phase space.

Our full physical transformer on manifold can be a pathway to developing powerful and potentially physical and unified reasoning frameworks.

Because the Neural Differential Manifold is equipped with Hamiltonian and HJB dynamics, our "physical transformer" can be viewed as a learned model of a complex, non-equilibrium dynamical system: internal states evolve on a latent manifold under Hamiltonian interactions, while FFN-like components and CTM-style temporal processing act as non-Hamiltonian baths that inject and dissipate energy. This geometry-plus-dynamics formulation is naturally suited for physical reasoning: physical systems are classically modeled as Hamiltonian flows on configuration manifolds, and control problems as HJB flows on those manifolds. In our framework, such physical flows coexist with graph-based episodic structure (GSW) and information-phase trajectories $(u_t, e_t)$, suggesting a pathway toward unified reasoning architectures in which physical, causal, and symbolic reasoning are different views of the same underlying manifold dynamics.

## Acknowledgements

The authors wish to acknowledge the use of AI-powered language models (ChatGPT) for assistance in improving the grammar, spelling, and readability of this manuscript.

## Author contributions

TX and ZH: Derive formulas and perform data analysis, LL: Problem formulation  formula derivation, MX: Design project and write paper.

## Competing interests

The authors declare no competing interests.